\documentclass[conference]{IEEEtran}
\IEEEoverridecommandlockouts
\usepackage{cite}
\usepackage{amsmath,amssymb,amsfonts}
\usepackage{algorithm}
\usepackage{algpseudocode}
\usepackage{graphicx}
\usepackage{textcomp}
\usepackage{xcolor}
\usepackage{microtype}
\usepackage{hyperref}
\usepackage{bm}
\usepackage{dsfont}
\usepackage{booktabs} % for professional tables
\usepackage{hhline}
\usepackage{array}
\usepackage{multirow}
\usepackage{tabularray}
\usepackage{csquotes}
\usepackage{mathtools}

\newtheorem{theorem}{Theorem}
\newtheorem{definition}{Definition}

\def\BibTeX{{\rm B\kern-.05em{\sc i\kern-.025em b}\kern-.08em
    T\kern-.1667em\lower.7ex\hbox{E}\kern-.125emX}}
\begin{document}

\title{Evaluating the Impact of Local Differential Privacy on Utility Loss via Influence Functions}

\author{\IEEEauthorblockN{Alycia N. Carey, Minh-Hao Van, Xintao Wu}
\IEEEauthorblockA{\textit{Department of Electrical Engineering and Computer Science} \\
\textit{University of Arkansas}\\
\{ancarey, haovan, xintaowu\}@uark.edu}
}

\maketitle

\begin{abstract}
How to properly set the privacy parameter in differential privacy (DP) has been an open question in DP research since it was first proposed in 2006. In this work, we demonstrate the ability of influence functions to offer insight into how a specific privacy parameter value will affect a model's test loss in the randomized response-based local DP setting. Our proposed method allows a data curator to select the privacy parameter best aligned with their allowed privacy-utility trade-off without requiring heavy computation such as extensive model retraining and data privatization. We consider multiple common randomization scenarios, such as performing randomized response over the features, and/or over the labels, as well as the more complex case of applying a class-dependent label noise correction method to offset the noise incurred by randomization. Further, we provide a detailed discussion over the computational complexity of our proposed approach inclusive of an empirical analysis. Through empirical evaluations we show that for both binary and multi-class settings, influence functions are able to approximate the true change in test loss that occurs when randomized response is applied over features and/or labels with small mean absolute error, especially in cases where noise correction methods are applied.
\end{abstract}


\section{Introduction}
Due to increased public awareness and demand of data privacy over the last decade, it has become common for companies like Google \cite{erlingsson2014rappor}, Apple \cite{apple-dp}, and Microsoft \cite{ding2017collecting} to integrate local differential privacy (LDP) \cite{dwork2006differential,kasiviswanathan2011can} into their data collection procedures. In LDP, users perturb their data locally using a randomization procedure before they are collected, thereby eliminating the reliance on a trustworthy aggregation server common in global DP applications. While several randomization procedures have been proposed \cite{10.1145/2746539.2746632,wang2017locally,erlingsson2014rappor} to increase the utility of the randomized data, most proposed works build upon the randomized response (RR) \cite{warner1965randomized} process proposed by Warner in 1965. RR was originally proposed as a technique to improve bias in collected survey responses and it ensures individual-level privacy by injecting plausible deniability into the collected data. 

In RR, the probability that a user answers truthfully (or reports their true value to an aggregation server) is set by the privacy parameter $\epsilon$, with smaller values of $\epsilon$ leading to greater privacy for the users. When deploying DP in real world settings, selecting $\epsilon$ that ensures a meaningful degree of privacy without significantly degrading the utility of the underlying system that uses the randomized data is a non-trivial task. It often requires heavy effort from skilled practitioners to choose the $\epsilon$ value that best balances this trade-off. If the practitioner chooses too small of an $\epsilon$, while the end-users' personal data are kept private, the general population-level statistics cannot be learned. On the other hand, if the practitioner chooses too large of an $\epsilon$, while being able to infer the wanted statistics, no meaningful notion of privacy is employed and it is as if the end-users' data were collected in the clear. Additionally, the application of DP is problem specific and there is minimal understanding of or guidance on how to choose $\epsilon$ for the specific task at hand \cite{dwork2019differential}. Further, while the paradigm of trading privacy for utility (and vice versa) is well understood, it is difficult to know a priori the effect that a certain $\epsilon$ will have on the utility of the model without actually perturbing the data and retraining the model. 

In this work, we take a step towards solving the problem of selecting $\epsilon$ in randomized response-based local differential privacy. We consider a modified LDP scenario in which a trusted data curator owns the original (non-privatized) data and wishes to use it to train a model that will be deployed publicly. Currently, selecting $\epsilon$ which gives the best privacy-utility trade-off requires perturbing the data and retraining the model for every $\epsilon$ under consideration -- which is a time and resource intensive process. We propose to overcome this prohibitive retraining requirement by leveraging influence functions from robust statistics to show the approximate change that would occur to a model's test loss under different $\epsilon$ values. Specifically, we present a solution that requires a fraction of the amount of retraining and does not require perturbing the original data. In short, we estimate the effect that a specific $\epsilon$ would have on the test loss of the model if it was actually used to perturb the data and the model was actually retrained. In this manner, we perform a \textit{`What\dots if\dots?'} analysis that answers questions analogous to \textit{`What would happen to the final model's performance if a certain group of training points were perturbed by a specific $\epsilon$?'} The focus on a group of training points may sound unreasonable, but we offer the following example to further motivate our work.  

\begin{displayquote}
\textit{Consider a company that routinely collects information from their end-users to train an in-house prediction model. The end-users consent to their data being used to train the model -- as long as the model is only used internally by the company. However, it is highly likely that at some point the company will decide to publish the model online, or sell it to another company, for profit. Some end-users wouldn't mind their data being used in this manner, but others would\footnote{ For example, smokers will be more concerned about their status as a smoker than non-smokers are due to stigma and impacts to health care if their status is revealed.} and would require the company to privatize their data using local differential privacy and retrain the model before its release. The company is therefore tasked with finding $\epsilon$ such that it provides sufficient privacy for the concerned end-users, but does not degrade the utility of the model.}
\end{displayquote}

There are several works similar in spirit to ours that aim to provide insight on selecting $\epsilon$ and the proper application of DP in practice. In \cite{lee2011much}, the authors state that while the privacy given by a certain $\epsilon$ has an intuitive theoretical interpretation, understanding the privacy of $\epsilon$ in practice is non-trivial. They additionally demonstrate the harm that can occur when $\epsilon$ is not carefully chosen to suit the problem at hand. Similarly, in \cite{mehner2021towards}, the authors offer an intuitive interpretation of $\epsilon$ based on quantifiable parameters, such as the number of end users and the sensitivity of the underlying system, to provide a more understandable statement on the the overall privacy risk posed by a certain $\epsilon$ value. Further, in \cite{hsu2014differential} the authors present a simple economic model for selecting $\epsilon$ by studying the impact of $\epsilon$ on both the data analyst and the prospective participants who contribute private data. We note, however, that all of these works focus on the central model of differential privacy, not local, and do not specifically consider the setting of machine learning. Furthermore, the previous work is quite different from ours and focuses on \textit{explaining the privacy} that a certain $\epsilon$ provides. While this work is crucial, we choose to focus on the equally important problem of \textit{explaining the effect} a certain $\epsilon$ has on the utility of a machine learning model when it is used to perturb a certain group of data points. To our knowledge, our work is the first to do so and analyze the impact of a certain $\epsilon$ on the test loss of a model in the randomized response-based LDP setting. 

Our contributions are as follows: (1) we present an approach for approximating the effect of perturbing a certain group of data points using randomized response-based $\epsilon$-LDP on a model's test loss based on influence functions. Our approach works in multiple common LDP scenarios, including randomizing the features, and/or randomizing the labels, as well as in the setting where label noise correction methods are applied. Additionally, our method allows for significant savings in computational costs when a large number of $\epsilon$ values and/or a large number of groups are involved in performing a \textit{`What... if...?'} analysis; (2) We perform empirical evaluations over two binary datasets and one multi-class dataset to show the ability of our method to accurately estimate the resulting change in test loss when different $\epsilon$ values and/or group sizes are used; and (3) we provide detailed timing analysis that shows using influence functions to approximate the true change in test lost saves computational time and resources compared to actually retraining the model for every $\epsilon$ and/or group construction under consideration.

The rest of the publication is as follows. We begin in Section \ref{sec:related} by introducing closely related work. Section \ref{sec:prelim} presents an overview of local differential privacy, randomized response, and influence functions. Building on Section \ref{sec:prelim}, in Section \ref{sec:methodology} we introduce both the generic and label specific methodologies to estimating the effect that a specific $\epsilon$ has on the final test loss of a model as well as perform a detailed time complexity analysis. Section \ref{sec:eval} details our experimentation and gives an empirical evaluation of our techniques. Finally, in Section \ref{sec:conclusion} we offer our concluding remarks and a discussion over our future work. 

\section{Related Work}
\label{sec:related}
\subsubsection{Local Differential Privacy}
Differential privacy (DP) is a formal notion of privacy that allows analysts to learn trends in sensitive data without revealing information specific to the individuals in the dataset \cite{dwork2009adifferential}. In central (also known as global) DP, we assume a trusted data curator collects all the sensitive data and ensures any queries over the data are made differentially private by adding noise proportional to the sensitivity of the query. Local differential privacy (LDP) \cite{kasiviswanathan2011can} eliminates the requirement of a trusted curator by requiring each user to perturb their own data before sending them to the server. Numerous research works have been published over LDP \cite{bebensee2019local,xiong2020comprehensive}, and LDP has been utilized in a myriad of different problem settings such as: data statistics and analysis \cite{wang2017locally}, graph neural networks \cite{sajadmanesh2021locally}, and federated learning \cite{geyer2017differentially}. LDP is also commonly used by the likes of Google \cite{erlingsson2014rappor}, Apple \cite{apple-dp}, and Microsoft \cite{ding2017collecting} to privatize collected information from their end users. Despite the real-world applications of LDP, it has been repeatedly noted that choosing a proper $\epsilon$ is complex and problem dependent. In \cite{garfinkel2018issues}, the US Census Bureau note that to pick a proper $\epsilon$ for US Census data release, they constructed a set of graphs showing the trade-off between $\epsilon$ and accuracy. However, constructing these graphs requires training multiple models, which is costly. Our approach based on influence functions eliminates the necessity of excessive model retraining and data randomization, enabling analysts to choose the proper $\epsilon$ faster and by using less compute resources. 

One common method to ensure LDP is randomized response. Randomized response was first proposed to improve bias in survey responses about sensitive issues like drug use \cite{chaudhuri2020randomized,warner1965randomized}. However, as techniques for differential privacy were being developed in the late 2000s, statisticians realized that randomized response inherently satisfied the requirements for being a differentially private algorithm. Since randomized response injects plausible deniability into the collected data, it naturally protects user's private data. In \cite{wang2016using}, the authors study how to enforce differential privacy by using randomized response in the data collection scenario and their work plays an important role in our formulations in Section \ref{sec:methodology}. 

\subsubsection{Influence Functions}
Influence functions are a product of influence analysis from the field of robust statistics \cite{cook1982residuals}. In influence analysis, small perturbations are introduced into the problem formulation (e.g., into the data or assumptions) and the resulting change in the outcome of the analysis is monitored \cite{cook1986assessment,cook1982residuals}. The idea of using influence functions to monitor change in a statistical model was extended in \cite{koh2017understanding} to show how a single training point influences the final machine learning model's parameters and/or the test loss of a single test point. This work was further extended in \cite{koh2019accuracy} in which the authors showed how influence functions can be used to estimate the influence that a group of training points has on the model parameters and/or the loss of a single test point. The methods formulated in these works serve as the basis for several extension works -- ours included. Specifically, influence functions have been used to estimate the influence of individual end-users in federated learning \cite{xue2021toward}, analyze the solutions produced by robust optimization \cite{deng2020interpreting}, perform out-of-distribution generalization analysis \cite{ye2021out}, and evaluate the fairness of a machine learning model \cite{li2022achieving,wang2022understanding}. Further, several works have been published over how to improve the approximation ability of  influence function-based methods \cite{bae2020influence,basu2020second,pruthi2020estimating}. 

While both influence functions and differential privacy have strong ties to robust statistics \cite{dwork2009adifferential,dwork2009bdifferential}, to our knowledge only one work has been published utilizing both methods \cite{kang2020differentially}. However, \cite{kang2020differentially} aims to improve the performance of differentially private empirical risk minimization by seeking out the training points with high influence and adding additional Gaussian noise to them. In contrast, our work focuses on showing the effect of $\epsilon$ on the model test loss and is the first to capitalize on the ability of influence functions to show the effect that a chosen $\epsilon$ has on the utility loss of a model. 

\section{Preliminaries}
\label{sec:prelim}
In this section, we present the required background information on local differential privacy and influence functions for understanding the discussions of Section \ref{sec:methodology}. We begin by detailing the notation used through the remainder of the paper. Let $\epsilon$ represent the privacy parameter in LDP and let $\mathbb{P}[\cdot]$ denote probability. Let $\mathcal{X}, \mathcal{Y}$ be the feature and label domain and $Z,Z_{te}$ represent the training and testing datasets. Let $z = (x\in\mathcal{X}, y\in\mathcal{Y}) \in Z$ represent one of $n$ training points and let  $z_{te} = (x_{te}\in\mathcal{X}, y_{te}\in\mathcal{Y}) \in Z_{te}$ represent one of $m$ testing points. We denote our model $h:\mathcal{X}\to\mathcal{Y}$, the model parameters by $\theta\in\Theta$, and use $\hat{\theta}$ to denote the optimal model parameters. We use $\ell(z,\theta)=\ell(h(x;\theta),y)$ to denote the loss function and $\mathcal{L}(Z, \theta) = \frac{1}{n}\sum_{i=1}^n\ell(z^i, \theta)$ to denote the empirical risk. The empirical risk minimizer is given by $\hat{\theta} = \arg\min_{\theta\in\Theta}\mathcal{L}(Z,\theta)$ and we assume that the empirical risk is twice-differentiable and strictly convex in $\theta$ \cite{koh2017understanding}.

\subsection{Local Differential Privacy}
As mentioned previously, local differential privacy allows an analyst to learn population statistics without violating the privacy of individuals. More formally, $\epsilon$-LDP is defined as follows:

\begin{definition}[$\epsilon$-LDP \cite{kasiviswanathan2011can}]
\label{def:ldp}
A randomized mechanism $\mathcal{M}$ satisfies $\epsilon$-local differential privacy if and only if for any pair of input values $r,r'$ in the domain of $\mathcal{M}$, and for any possible output $o\in\text{Range}(\mathcal{M})$, it holds:
\begin{equation*}\small
    \mathbb{P}[\mathcal{M}(r) = o] \leq e^{\epsilon}\cdot\mathbb{P}[\mathcal{M}(r') = o]
\end{equation*}
\end{definition}

Definition \ref{def:ldp} states that the probability of outputting $o$ on record $r$ is at most $e^\epsilon$ times the probability of outputting $o$ on record $r'$.

\subsubsection{Randomized Response}
One popular method used to implement LDP is randomized response. Let $u$ be a private variable that can take one of $C$ values. We can formalize the randomize response process as a $C\times C$ distortion matrix $\mathrm{\textbf{P}}=(p_{uv})_{C\times C}$ where $p_{uv} = \mathbb{P}[v | u] \in (0,1)$ denotes the probability that the output of the randomized response process is $v\in \{1,\dots,C\}$ when the real attribute value is $u\in \{1,\dots,C\}$. Note that the entries of the distortion matrix are probabilities, and therefore the sum of the probabilities of each row is 1 \cite{wang2016using}. Further, $\mathrm{\textbf{P}}$ can be altered to achieve both optimal utility and $\epsilon$-DP by setting as follows \cite{wang2016using}:
\begin{equation}\small
    \label{eq:pert-mat}
    p_{uv}=
        \begin{cases}
            \frac{e^\epsilon}{C-1+e^\epsilon} & \text{if } u = v\\
            \frac{1}{C-1+e^\epsilon} & \text{if } u \neq v
    \end{cases}
\end{equation}

If we have a dataset that was collected using randomized response, then using the distortion matrix $\mathrm{\textbf{P}}$ that was used during data collection, we can estimate the true population distribution from the noisy collected data. Let $\bm{\pi} = \{\pi_1, \dots, \pi_C\}$ be the true (to be estimated) proportion of the values in the original population and let $\bm{\lambda} =\{\lambda_1, \dots, \lambda_C\}$ be the observed proportion of the values in the collected noisy dataset. Using the relationship $\bm{\pi} \approx \mathrm{\textbf{P}}^{-1}\bm{\lambda}$ we can estimate the true underlying population $\bm{\pi}$ based on the observed values in the collected noisy dataset $\bm{\lambda}$. We note that in our setting a trustworthy server has all the original statistics $\bm{\pi}$. Therefore, to estimate the values of $\bm{\lambda}$ without actually perturbing the data, we can calculate $\bm{\lambda} \approx \mathrm{\textbf{P}}\bm{\pi}$. We discuss this idea further in Section \ref{sec:methodology}.

\subsection{Influence Functions}
In \cite{koh2017understanding}, the authors propose the use of influence functions to study machine learning models through the lens of their training data. Specifically, they show that the influence a single training point $z = (x, y)$ has on the model parameters can be calculated without actually removing $z$ from the training set and retraining the model on the resulting dataset. They instead simulate the removal of $z$ by upweighting it by a small value $\frac{1}{n}$ (where $n$ is the total number of training points). They then calculate the influence the training point has on the model parameters as: 
\begin{equation}\small
    \label{eq:up_param}
    \mathcal{I}_{up,params}(z) = -H_{\hat{\theta}}^{-1}\nabla_{\theta}\ell(z,\hat{\theta})
\end{equation}
where $H_{\hat{\theta}}^{-1}$ is the inverse Hessian matrix:
\begin{equation}\small
    \label{eq:hessian}
    H_{\hat{\theta}}^{-1} = \Bigg(\frac{1}{n} \sum_{i=1}^n\nabla_{\theta}^2\ell(z^i,\hat{\theta})\Bigg)^{-1}
\end{equation}
Note that the inverse Hessian matrix can be calculated explicitly as defined in Eq. \ref{eq:hessian} or efficiently estimated using the conjugate gradient or stochastic estimation approaches \cite{koh2017understanding}. We discuss the implications of this further in Section \ref{sec:method-time}.

Eq. \ref{eq:up_param} is obtained by performing a quadratic expansion around the optimal parameters $\hat{\theta}$ which gives an approximation of the function locally using information about the steepness (the gradient) and the curvature (the Hessian). Eq. \ref{eq:up_param} can be used to approximate the parameters that would be obtained if $z$ was actually removed from the dataset and the model was retrained as:
\begin{equation}\small
    \hat{\theta}_{-z} \approx \hat{\theta} - \frac{1}{n}\mathcal{I}_{up,params}(\hat{\theta}, z)
\end{equation}

In \cite{koh2017understanding}, the authors further extend Eq. \ref{eq:up_param} to show the influence a training instance $z$ has on the loss of a test instance $z_{te}$:
\begin{equation}\small
    \label{eq:up_loss}
        \mathcal{I}_{up,loss}(z, z_{te}) = -\nabla_{\theta}\ell(z_{te},\hat{\theta})^TH_{\hat{\theta}}^{-1}\nabla_{\theta}\ell(z,\hat{\theta})
\end{equation}
where $\nabla_{\theta}\ell(z_{te},\hat{\theta})$ is the gradient of the test instance w.r.t. the optimal model parameters and $\frac{1}{n}\mathcal{I}_{up,loss}(z,z_{te})$ gives the approximate change in loss for test point $z_{te}$. The authors of \cite{koh2017understanding} also consider the effect that perturbing a training point has on the parameters/loss of a test point. Consider a training point $z$ and its perturbed\footnote{Here, we consider a general perturbation that can be discrete or continuous. We are not specifically considering randomized response perturbations.} value $z_{\beta}$. Let: 
\begin{equation}\small
    \hat{\theta}_{z_{\beta}, -z} = \arg\min_{\theta\in\Theta}\mathcal{L}(Z,\theta)+\frac{1}{n}\ell(z_{\beta}, \theta) - \frac{1}{n}\ell(z, \theta)
\end{equation}
be the empirical risk minimizer on the training points with $z_{\beta}$ in place of $z$. The approximate effect that changing $z$ to $z_{\beta}$ has on the model parameters can be computed as:
\begin{equation}\small
\label{eq:pert-params}
    \mathcal{I}_{pert, params}(z_{\beta},-z)  = -H_{\hat{\theta}}^{-1}\Big(\nabla_{\theta}\big(\ell(z_{\beta}, \hat{\theta}) - \ell(z,\hat{\theta})\big)\Big)\\
\end{equation}
which can then use to approximate the new parameters as:
\begin{equation}\small
    \label{eq:pert-param}
    \hat{\theta}_{z_{\beta},-z}  \approx \hat{\theta}+\frac{1}{n}\mathcal{I}_{pert,params}(z_{\beta},-z)
\end{equation}

As before, the authors of \cite{koh2017understanding} extend Eq. \ref{eq:pert-params} to show the approximate effect that perturbing $z$ to $z_{\beta}$ has on the loss of a test point $z_{te}$: 
\begin{equation}\small
    \label{eq:pert-loss}
    \begin{aligned}
    \mathcal{I}_{pert,loss}& (z_{\beta}, -z, z_{te})  = \\
    &-\nabla_{\theta}\ell(z_{te},\hat{\theta})^TH_{\hat{\theta}}^{-1}\Big(\nabla_{\theta}\big(\ell(z_{\beta}, \hat{\theta}) - \ell(z,\hat{\theta})\big)\Big)
    \end{aligned}
\end{equation}
and $\frac{1}{n}\mathcal{I}_{pert,loss}(z_{\beta},-z,z_{te})$ is the approximate change in loss for test point $z_{te}$. 

\section{Methodology}
\label{sec:methodology}
We now detail our approach based on influence functions for estimating the effect that perturbing a group of training points $S \subset Z$ under randomized response-based $\epsilon$-LDP would have on the test loss of a model if the perturbation was actually performed and the model was retrained. We begin by giving a formulation of Eq. \ref{eq:pert-loss} in the \textit{Group-to-Group} setting which is followed by the presentation of the general influence formula for approximating the effect of applying randomized response-based $\epsilon$-LDP on the features, labels, or features and labels. We then focus specifically on the label perturbation scenario and show how our proposed formulation can be altered to approximate the effect of applying a noise correction method after randomized response-based $\epsilon$-LDP is performed. Finally, we offer an analysis of the computational complexity of our approach compared to the na\"{i}ve approach of model retraining as well as a discussion over the ability of our method to extend to other LDP protocols beyond randomized response.

\subsection{Influence of $\epsilon$-LDP on Model Utility Loss}
\subsubsection{Group-to-Group Influence}

\begin{algorithm}[ht!]\small
    \caption{Approximate Effect of $\epsilon$}
    \label{alg:eq10}
    \begin{algorithmic}[1]
        \renewcommand{\algorithmicrequire}{\textbf{Input:}}
        \Require $Z$, $G$, $G_{te}$, $\epsilon$
        \State $\hat{\theta} = \arg\min_\theta\frac{1}{n}\sum^n_{i=1}\ell(h(x^i;\theta), y^i)$ \Comment{Train model}
        \State $H_{\hat{\theta}}^{-1} = \big(\frac{1}{n}\sum^n_{i=1}\nabla^2_\theta\ell(z^i,\hat{\theta})\big)^{-1}$ \Comment{Compute Hessian}
        \State $\mathcal{L}_{te} = 0$
        \For{$z_{te} \in G_{te}$} \Comment{Calculate test loss}
            \State $\mathcal{L}_{te} = \mathcal{L}_{te} + \ell(z_{te},\hat{\theta})$
        \EndFor
        \State $\eta = -\nabla_\theta \frac{\mathcal{L}_{te}}{m}$ \Comment{Gradient of average test loss}
        \State $\mathcal{L}$ = 0
        \For{$z \in G$} \Comment{Calculate loss difference}
            \State $z_\beta = \text{RR}(z, \epsilon)$ 
            \State $\mathcal{L} = \mathcal{L} + (\ell(z_\beta, \hat{\theta}) - \ell(z, \hat{\theta}))$
        \EndFor
        \State $\gamma = \nabla_\theta \mathcal{L}$ \Comment{Gradient of total loss difference}
        \State $\mathcal{I}_{pert,loss} = \eta\cdot H_{\hat{\theta}}^{-1}\cdot\gamma$\\
        \Return $\mathcal{I}_{pert,loss}$
    \end{algorithmic}
\end{algorithm}
In \cite{koh2019accuracy}, the authors showed that influences are additive with respect to a single test point. For example, given a group of training points $G\subset Z$ and a single test point $z_{te}$:
\begin{equation}\scriptsize
\label{eq:group-to-single}
\begin{aligned}
    \mathcal{I}_{pert,loss}&(G_\beta, -G, z_{te})  \\
    &= \sum_{i\in G}\mathcal{I}_{pert,loss}(z_\beta^i, -z^i, z_{te})\\
    &=-\nabla_{\theta}\ell(z_{te},\hat{\theta})^TH_{\hat{\theta}}^{-1}\Bigg(\nabla_{\theta}\sum_{i\in G}\big(\ell(z^i_{\beta}, \hat{\theta}) - \ell(z^i,\hat{\theta})\big)\Bigg)
\end{aligned}
\end{equation}
where $G_\beta = \{z_\beta^i\}_{i=1}^{|G|}$, $G = \{z^i\}_{i=1}^{|G|}$. However, in this work we consider the influence that a group of training points has on the loss of a group of test points. To achieve this goal, we begin by extending Eq. \ref{eq:group-to-single} to calculate the influence on the loss of a group of test points $G_{te}\subseteq Z_{te}$:
\begin{equation}\scriptsize
\label{eq:group-to-group-gen}
\begin{aligned}
    \mathcal{I}^{\text{G2G}}_{pert,loss}&(G_\beta, -G, G_{te}) = \sum_{j\in G_{te}}\sum_{i\in G}\mathcal{I}_{pert,loss}(z_\beta^i, -z^i, z^j_{te})\\
    &=-\nabla_{\theta}\mathcal{L}(G_{te},\hat{\theta})^TH_{\hat{\theta}}^{-1}\Bigg(\nabla_{\theta}\sum_{i\in G}\big(\ell(z^i_{\beta}, \hat{\theta}) - \ell(z^i,\hat{\theta})\big)\Bigg)
\end{aligned}
\end{equation}
where $G_{te} = \{z_{te}^j\}_{j=1}^{|G_{te}|}$. 

We present the process of calculating $\mathcal{I}^{\text{G2G}}_{pert,loss}$ in Alg. \ref{alg:eq10}. In the first line, we train the model $h$ over the training data $Z$ to produce the optimal parameters $\hat{\theta}$ -- which is used as the baseline model in the calculation of the influence function. In line 2, we begin to calculate $\mathcal{I}^{\text{G2G}}_{pert,loss}$ by first computing the inverse Hessian over the original training data. We note that the inverse Hessian only needs to be computed once even if multiple different $\epsilon$ values and group constructions $G$ are being tested. In lines 3-7, we compute the gradient of the loss over the entire test group $G_{te}$. As with calculating the inverse Hessian matrix, as long as the construction of $G_{te}$ does not change, this line only has to be performed once. Lines 8-12 compute the aggregated differences between the loss when the data point $z \in G$ is perturbed to $z_\beta$ using randomized response-based $\epsilon$-LDP (line 10) and the original loss on point $z$. Lines 13 and 14 finish out the computation by first taking the gradient of the aggregated loss computed in lines 8-12 and then multiplying the results of lines 2 (the inverse Hessian), 7 (the gradient of the test loss), and 13 (the gradient of the aggregate loss). In Alg. \ref{alg:eq10}, randomization is actually performed on the training point $z$. However, in the next subsection we will show how $\mathcal{I}^{\text{G2G}}_{pert,loss}$ can be modified to obtain the estimated effect of randomization without actually having to perturb the training point. We also note that $z^i_\beta$ can be constructed three different ways -- $z^i_\beta=(x^i_\beta, y^i)$, $z^i_\beta = (x^i, y^i_\beta)$, and $z^i_\beta = (x^i_\beta, y^i_\beta)$ -- without altering the formulation of Eqs. \ref{eq:pert-param} - \ref{eq:group-to-group-gen}. This idea is utilized in the next subsection to formulate our general influence formula to estimate the effect of randomized response $\mathcal{I}^{\text{RR}}_{pert,loss}$.

\subsubsection{$\mathcal{I}^{\text{RR}}_{pert,loss}$}
In this subsection, we make the generalization of $\mathcal{A} = \mathcal{X}\cup\mathcal{Y}$. In other words, instead of considering the features and label domain as separate (e.g., $z=(x,y)$) we instead think of them as a combined attribute domain (e.g., $z=(a)$). We place a 
further restrictions on $\mathcal{A}$ such that $\mathcal{A} = \{A_1, \dots, A_T\}$ is the set of attributes where each $A_t=\{a_1, \dots, a_{d_t}\}$ has $d_t$ mutually exclusive and exhaustive categories. In our scenario, we consider that a subset of attributes $\mathcal{F}\subseteq \mathcal{A}$ are perturbed via randomized response (meaning that both the features and the label could possibly be perturbed). Let $\mathcal{F}_\times$ represent the Cartesian product of the elements of $\mathcal{F}$. Let each feature $A_f\in\mathcal{F}$ be associated with a randomized response distortion matrix $\text{\textbf{P}}^f=\{p_{uv}\}_{d_f \times d_f}$ where:
\begin{equation}\small
\label{eq:multiattribute-p}
    p_{uv} = \begin{cases}
        \frac{\epsilon_f}{d_f-1+e^{\epsilon_f}} & \text{if } u = v\\
        \frac{1}{d_f-1+e^{\epsilon_f}} & \text{if } u \neq v
    \end{cases}
\end{equation}
and $\epsilon_f$ represents the privacy parameter for attribute $A_f$. 

Let $\mathcal{V} = \mathcal{A}-\mathcal{F}$. Since $\mathcal{V}\cap\mathcal{F} =\emptyset$, we can write $z=(a)$ as $z=(\alpha\cup\delta)$ where $\alpha=\{a_t \in \mathcal{F}\}$ and $\delta=\{a_t\in \mathcal{V}\}$. For example, let $\mathcal{A}=\{gender, race, age, income\}$ and let $\mathcal{F} = \{gender, income\}$. If $gender=\{m,f\}$ and $income=\{0,1\}$, then, in this setting, $\mathcal{F}_\times=\{(m,0), (m,1), (f,0), (f,1)\}$. If we have a training point $z=(a=\{f,B,45,1\})$, then we can rewrite it as $z=(\alpha\cup \delta)$ where $\alpha=\{f,1\}$ and $\delta=\{B, 45\}$.

When computing the influence function, we only consider cases where the attributes that are modified under randomized response are not the same as the original attribute combination. I.e., we want $f_\times\in\mathcal{F}_\times- \alpha$. Note that according to Eq. \ref{eq:multiattribute-p} we have probability $\frac{1}{d_f-1+e^{\epsilon_f}}$ of changing the one attribute value $a_t\in\mathcal{A}$ to $a_{t_\beta}$. In order to correctly calculate the probability of $a$ being perturbed to another element in $\mathcal{F}_\times-\alpha$, we have to consider the probability of all elements in $\mathcal{F}_\times-\alpha$ being the outcome of the randomized response process. This combined probability can be calculated as $ 1 - \prod_{f\in\mathcal{F}}\frac{e^{\epsilon_f}}{d_f - 1 +  e^{\epsilon_f}}$.

Using these ideas, we can write the group to group influence of applying randomized response ($\mathcal{I}_{pert,loss}^{\text{RR}}$) as:
\begin{equation}\scriptsize
\label{eq:general-inf-rr}
\begin{aligned}
    &\mathcal{I}_{pert,loss}^{\text{RR}}(S, S_{te}, \bm{\epsilon}) =-\nabla_{\theta}\mathcal{L}(S_{te};\hat{\theta})H_{\hat{\theta}}^{-1}\cdot\\ &\nabla_\theta\Bigg(\bigg(1-\prod_{f\in\mathcal{F}}\frac{e^{\epsilon_f}}{d_f-1+e^{\epsilon_f}}\bigg)\sum_{i\in S}\sum_{f_{\times}\in \mathcal{F}_\times- \alpha^i}\big(\ell(f_\times,\hat{\theta})-\ell(z^i;\hat{\theta})\big)\Bigg)
\end{aligned}
\end{equation}
where $\bm{\epsilon} = \{\epsilon_f\}_{f\in\mathcal{F}}$ and $\frac{1}{n}\mathcal{I}_{pert,loss}^{\text{RR}}$ gives the estimated change in test loss.

\subsection{Influence of $\epsilon$-LDP Labels on Model Utility Loss}
We now use Eq. \ref{eq:general-inf-rr} as the base of our formulation for estimating the effect that perturbing only the training labels using randomized response-based $\epsilon$-LDP has on the model's final test loss. We note that in this section we detail the formulation for approximating the effect of label perturbation (and not feature or feature and label perturbation) to give a foundation upon which we can build our construction of approximating the effect of applying randomized response-based $\epsilon$-LDP on the labels with class-dependent label noise correction in the next section. Recall the distortion matrix $\mathrm{\textbf{P}}$ as defined in Section \ref{sec:prelim}. The probability that a training point $z^i = (x^i, y^i)$ is to perturbed to $z^i_\beta=(x^i, c)$ is $\frac{1}{C-1+e^{\epsilon}}$. Using this probability, and the idea that the influence function is only calculated over modified points (i.e., $c\neq y^i$), we can estimate the effect that applying $\epsilon$-LDP to group $S \subset Z$ would have on the loss of a group of test points $S_{te} \subseteq Z_{te}$ as:
\begin{equation}\scriptsize
    \label{eq:group-to-group}
    \begin{aligned}
  \mathcal{I}_{pert,loss}^{\text{RR-L}}(&S, S_{te}, \epsilon)=
    -\nabla_{\theta}\mathcal{L}(S_{te},\hat{\theta})^TH_{\hat{\theta}}^{-1}\cdot\\
    &\nabla_{\theta}\Bigg(\frac{1}{C-1+e^{\epsilon}}\sum_{i\in S}\sum_{c\in C \atop c\neq y^i}\ell(h(x^i;\hat{\theta}), c) \ell(h(x^i;\hat{\theta}), y^i)\Bigg)
    \end{aligned}
\end{equation}
where $\frac{1}{n}\mathcal{I}_{pert,loss}^{\text{RR-L}}(S, S_{te}, \epsilon)$ gives the estimated change in model test loss when $\epsilon$-LDP is applied to the labels.

\subsubsection{Forward Loss Correction}
\label{sec:flc}
When randomized response is used to perturb labels, it is common that a noise correction procedure is used to counteract the injected noise from randomization. Once such noise correction procedure is Forward Loss Correction (FLC) \cite{patrini2017making}. FLC is an approach to train machine learning models robust to class-dependent label noise (not necessarily noise crafted by randomized response). They note that a model learned without using loss correction would result in the model being tailored to predict noisy labels instead of the actual labels. To perform FLC, the authors correct the model predictions using a probability matrix that defines the noisy data distribution (in our case, the probability matrix is the distortion matrix $\mathrm{\textbf{P}}$) before calculating the loss between the model prediction and the noisy labels $\tilde{y} \in \tilde{\mathcal{Y}}$. FLC is defined as:
\begin{equation}\scriptsize
    \label{eq:flc-orig}
    \mathcal{L}^{\text{FLC}}(\tilde{Z}, \theta) = \frac{1}{n}\sum_{i=1}^n\ell(\mathrm{\textbf{P}}^Th(x^i;\theta), \tilde{y}^i)
\end{equation}
where $\tilde{Z}=\{\tilde{z}^i=(x^i,\tilde{y}^i)\}^n_{i=1}$ and $\ell$ is a proper composite loss\footnote{A proper loss is a loss function that both predicts the binary classification label as well as provides an estimate of the probability that an example will have positive label. Proper losses are called proper composite losses when a link function is used to map the output of the predictor to the interval $[0,1]$ in order for the output to be interpreted as a probability \cite{reid2010composite}.} \cite{reid2010composite} such as cross-entropy or square loss. Here, we will slightly abuse our notation $h(x;\theta)$ in order to explicitly show that $\ell$ is a proper composite loss. Thus far, we have considered $h$ to map $\mathcal{X}\to\mathcal{Y}$. In other words, the output of $h(x;\theta)$ is the predicted class label. However, for this section we consider $h(x;\theta)=\mathbb{P}(y\mid x)$. In other words, the output of $h$ is the predicted probability of the class being $y$ when the input is $x$. Using this notation, we can rewrite Eq. \ref{eq:flc-orig} as:
\begin{equation}\scriptsize
    \label{eq:flc}
    \mathcal{L}^{\text{FLC}}_\psi(\tilde{Z}, \theta) = \frac{1}{n}\sum_{i=1}^n\ell(\mathrm{\textbf{P}}^T\psi^{-1}(h(x^i;\theta)), \tilde{y}^i)
\end{equation}
Here, $\psi$ is the link function associated with a particular proper loss. For example, softmax is the inverse link function for cross-entropy. When FLC is applied while minimizing a proper composite loss function, \cite{patrini2017making} notes that the minimizer of the corrected loss under the noisy distribution is the same as the minimizer of the original loss under the clean distribution:
\begin{equation}\scriptsize
    \label{eq:flc-min}
    \hat{\theta}=\arg\min_{\theta\in\Theta}\mathcal{L}_\psi^{\text{FLC}}(\tilde{Z}, \theta) = \arg\min_{\theta\in\Theta}\mathcal{L}_\psi(Z, \theta)
\end{equation}
In other words, the learned model will make correct predictions on future non-randomized test data. For brevity, we refer readers to \cite{patrini2017making} for an in-depth discussion of FLC.

Our scenario varies slightly from \cite{patrini2017making} in that they assume all labels have been randomly perturbed, while we assume only a known group are. Therefore, if we use Eq. \ref{eq:flc} as given in our approach, we will overcompensate for the noise caused by applying randomized response to only group $S$. To solve this issue, we propose Theorem \ref{thm:flc-adjusted}.

\begin{theorem}
\label{thm:flc-adjusted}
Suppose that the distortion matrix $\mathrm{\textbf{P}}$ is non-singular. Define the adjusted forward loss correction as:
\begin{equation}\scriptsize
\label{eq:thm-loss}
\begin{aligned}
    &\mathcal{L}^{\text{FLC}}_\psi(S\cup R, \theta) = \\
    & \quad\frac{1}{n}\Bigg(\sum_{i\in R}\ell(\psi^{-1}(h(x^i;\theta)), y^i) + \sum_{j\in S}\ell(\mathrm{\textbf{P}}^T\psi^{-1}(h(x^j;\theta)), \tilde{y}^j)\Bigg)
\end{aligned}
\end{equation}
where $S\subset Z$ is the group of perturbed points and $R = Z\backslash S$. Then, the minimizer of the corrected loss under both the noisy and clean distribution is the same as the minimizer of the original loss under the entire clean distribution: 
\begin{equation}\scriptsize
\hat{\theta}=\arg\min_{\theta\in\Theta}\mathcal{L}^{\text{FLC}}_\psi(S\cup R, \theta) = \arg\min_{\theta\in\Theta}\mathcal{L}_\psi(Z, \theta)
\end{equation}

\end{theorem}
\begin{IEEEproof}
Assume there are $K$ disjoint subsets of the training dataset $Z = G^1 \cap G^2 \cap \dots \cap G^K$ each of which is defined by a $C\times C$ perturbation matrix $\mathrm{\textbf{P}^{k}}$ where $C = \{1,\dots,C\}$ represent the possible labels. Assume that the make up of each subgroup $G^i$, $\bm{\pi}^i = \{\pi^i_1, \cdots, \pi^i_C\}$, is known. Then:
\begin{equation}\scriptsize
\label{eq:proof}
\mathrm{\textbf{P}}^{1}\bm{\pi}^1+ \mathrm{\textbf{P}}^{2}\bm{\pi}^2 + \dots + \mathrm{\textbf{P}}^{K}\bm{\pi}^K = \mathrm{\textbf{P}}\bm{\pi}
\end{equation}
where $\mathrm{\textbf{P}}$ is a matrix with $p_{uv}=\sum_{i\in K}\frac{\pi_u^i}{\pi_u}p_{uv}^i$ and $\bm{\pi}$ is a vector with $\pi_c=\sum_{i\in K}\pi_c^i$. Eq. \ref{eq:proof} shows that we can write the $K$ different group-level data distributions as one population level distribution. Therefore, the loss function in Eq. \ref{eq:thm-loss} reduces to Eq. \ref{eq:flc} and the proof provided in \cite{patrini2017making} follows directly. In short, \cite{patrini2017making} proves Eq. \ref{eq:flc-min} by showing that by combining $\mathrm{\textbf{P}}^T$ with $\psi^{-1}$ (specifically $\phi^{-1}=\psi^{-1}\circ\mathrm{\textbf{P}}^T$) a new link function $\phi$ is formed and that the following holds:
\begin{equation}\scriptsize
    \begin{aligned}
            \mathcal{L}_{\phi}(\tilde{\mathcal{Z}},\theta) & = \frac{1}{n}\sum^n_{i=1}\ell(\phi(h(x^i,\theta)), \tilde{y}^i) \\ 
            & = \frac{1}{n}\sum^n_{i=1}\ell(\psi((\mathrm{\textbf{P}^{-1}})^Th(x^i,\theta)), \tilde{y}^i) \\
            & = \frac{1}{n}\sum^n_{i=1}\ell(\psi(h(x^i,\theta)), y^i)
    \end{aligned}
\end{equation}
\end{IEEEproof}

Theorem \ref{thm:flc-adjusted} gives intuition on how we can incorporate the distortion matrix $\mathrm{\textbf{P}}$ into the influence function in order to show how implementing FLC to correct noise from $\epsilon$-LDP affects the model's final test loss. Specifically, since influence functions only consider training instances of interest (i.e., those removed/modified/perturbed), we can modify Eq. \ref{eq:group-to-group} to consider FLC without over correcting the loss:
\begin{equation}\scriptsize
    \label{eq:group-to-group-FLC}
    \begin{aligned}
    &\mathcal{I}_{pert,loss}^{\text{RR-FLC}}(S, S_{te}, \epsilon)=-\nabla_{\theta}\mathcal{L}(S_{te},\hat{\theta})^TH_{\hat{\theta}}^{-1}\cdot\\
    &\qquad\nabla_{\theta}\Bigg(\frac{1}{C-1+e^{\epsilon}}\sum_{i\in S}\sum_{c\in C \atop c\neq y^i}\ell(\mathrm{\textbf{P}}^Th(x^i;\hat{\theta}), c)- \ell(h(x^i;\hat{\theta}), y^i)\Bigg)
    \end{aligned}
\end{equation}
where $\frac{1}{n}\mathcal{I}_{pert,loss}^{\text{RR-FLC}}(S, S_{te}, \epsilon)$ gives the estimated change in model test loss when the effect of using FLC to correct the noise of $\epsilon$-LDP is simulated. 

\subsection{Discussion}

\begin{figure*}[t!]
    \centering
    \includegraphics[width=.25\textwidth]{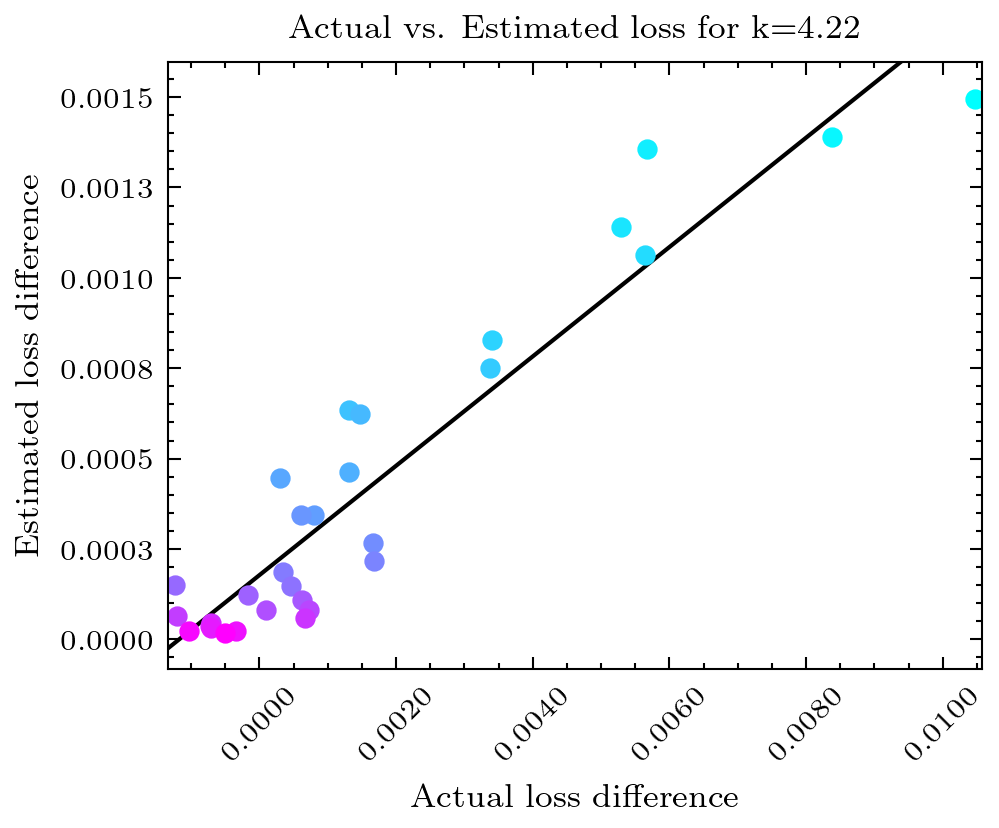}\hfill
    \includegraphics[width=.25\textwidth]{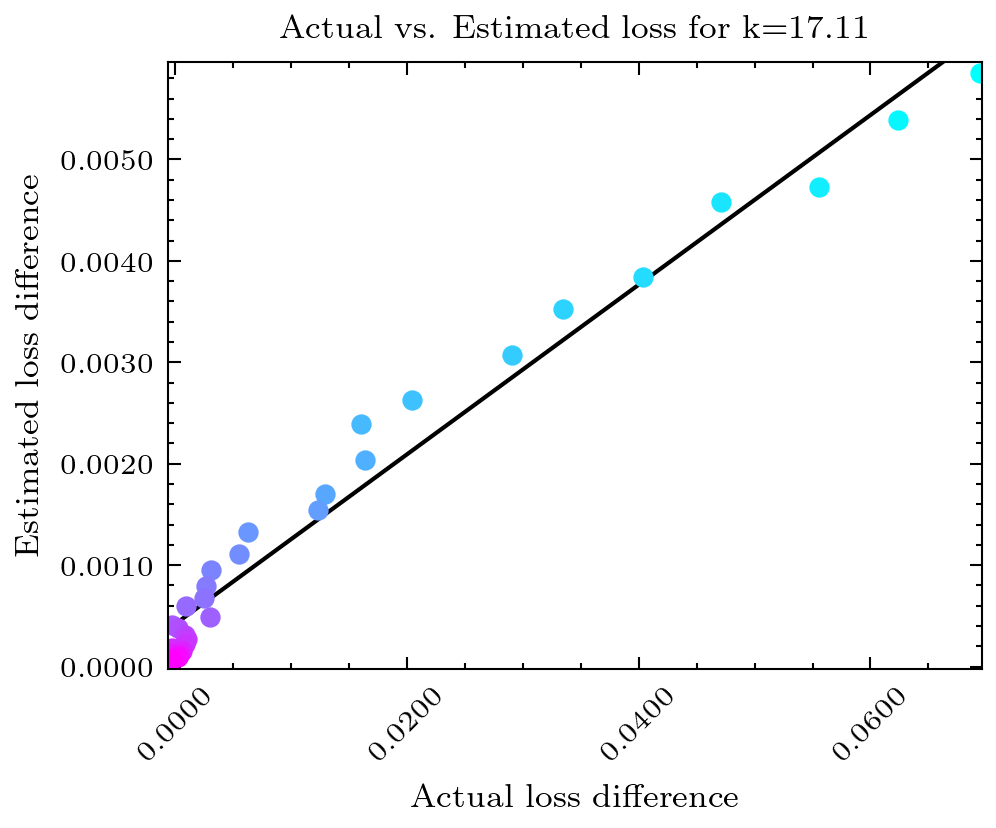}\hfill
    \includegraphics[width=.25\textwidth]{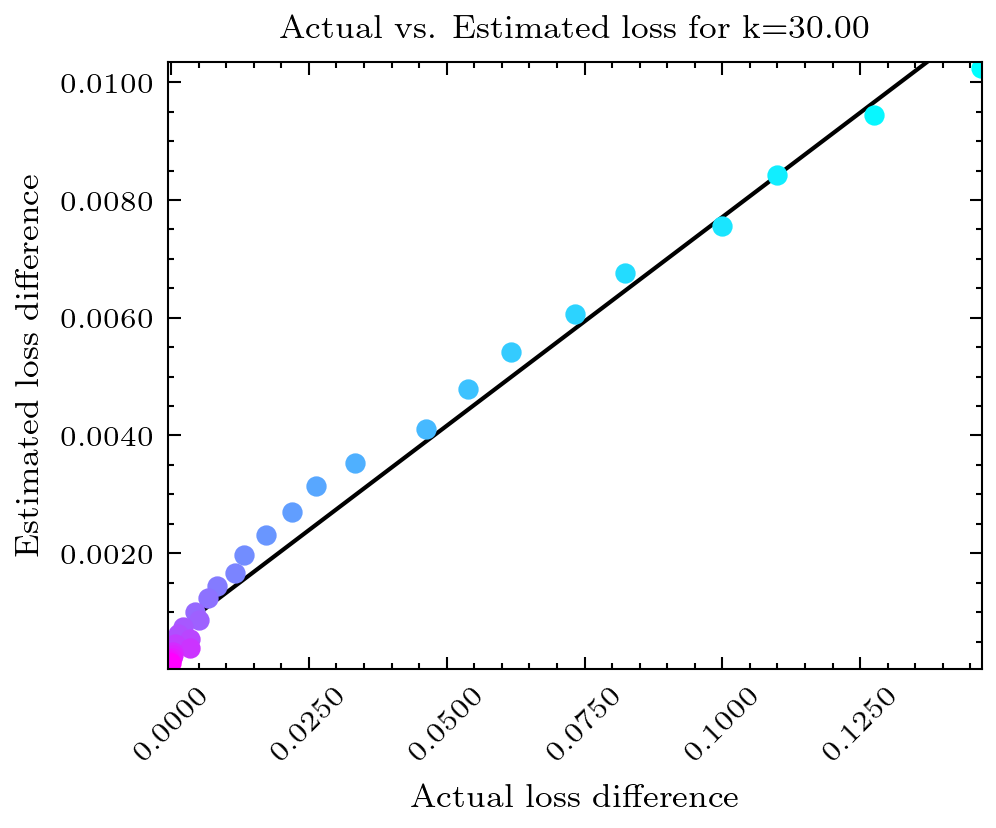}
    \includegraphics[width=.25\textwidth]{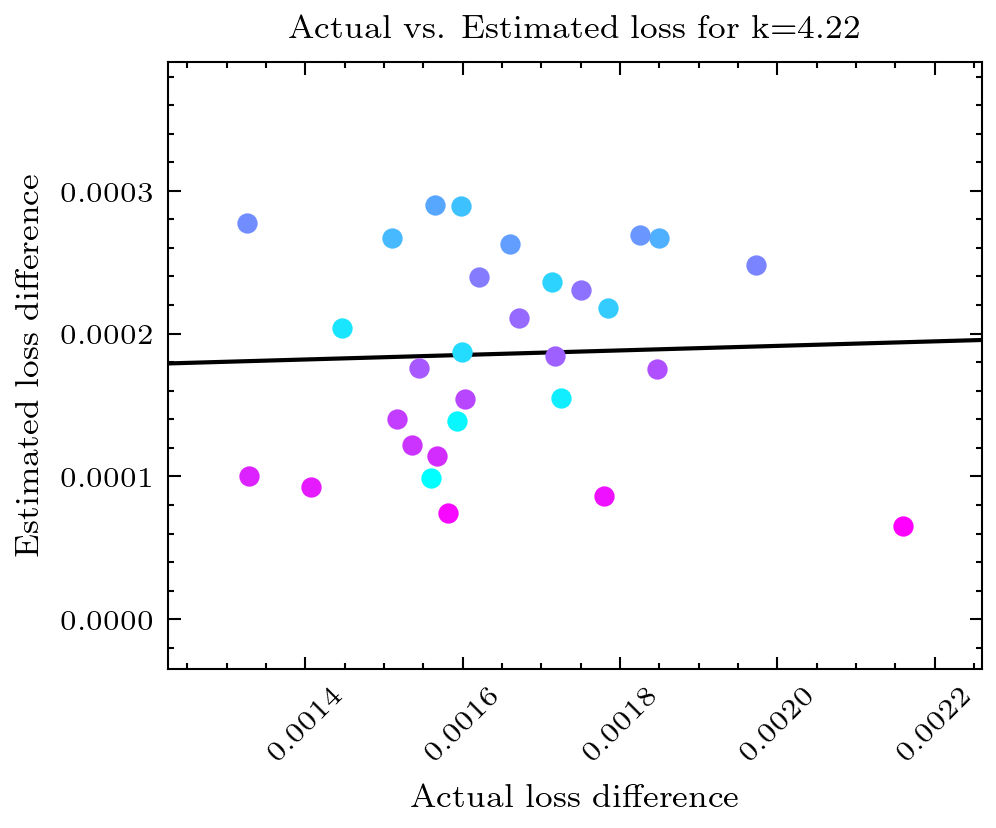}\hfill
    \includegraphics[width=.25\textwidth]{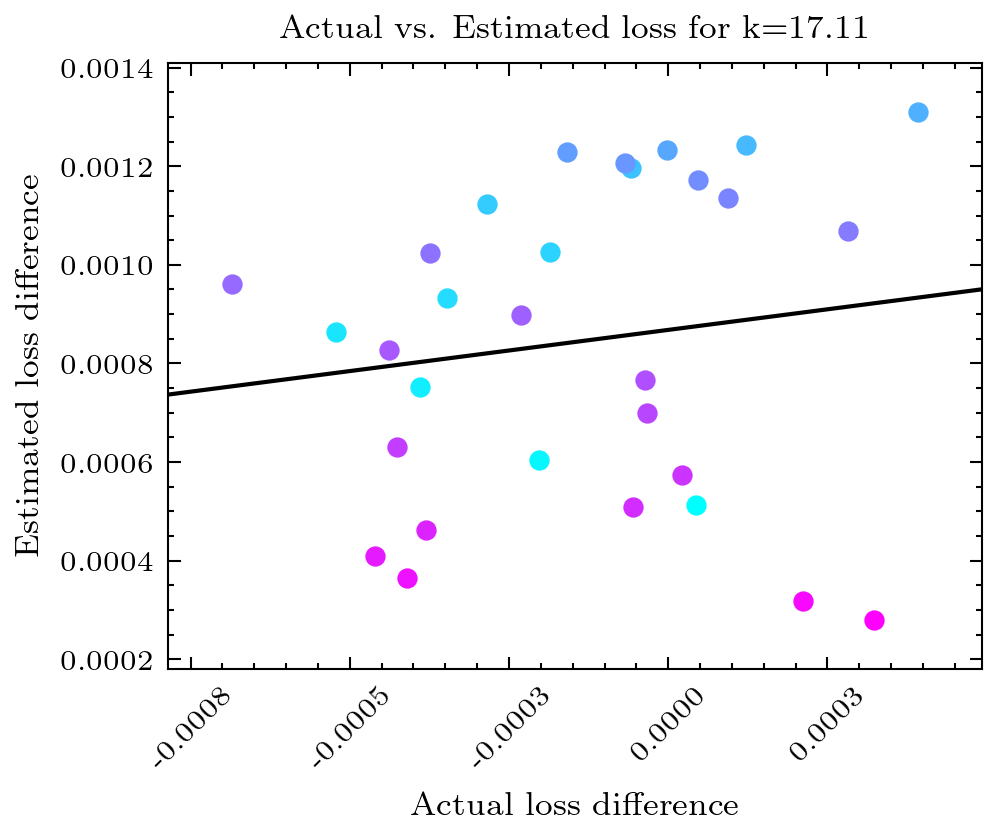}\hfill
    \includegraphics[width=.25\textwidth]{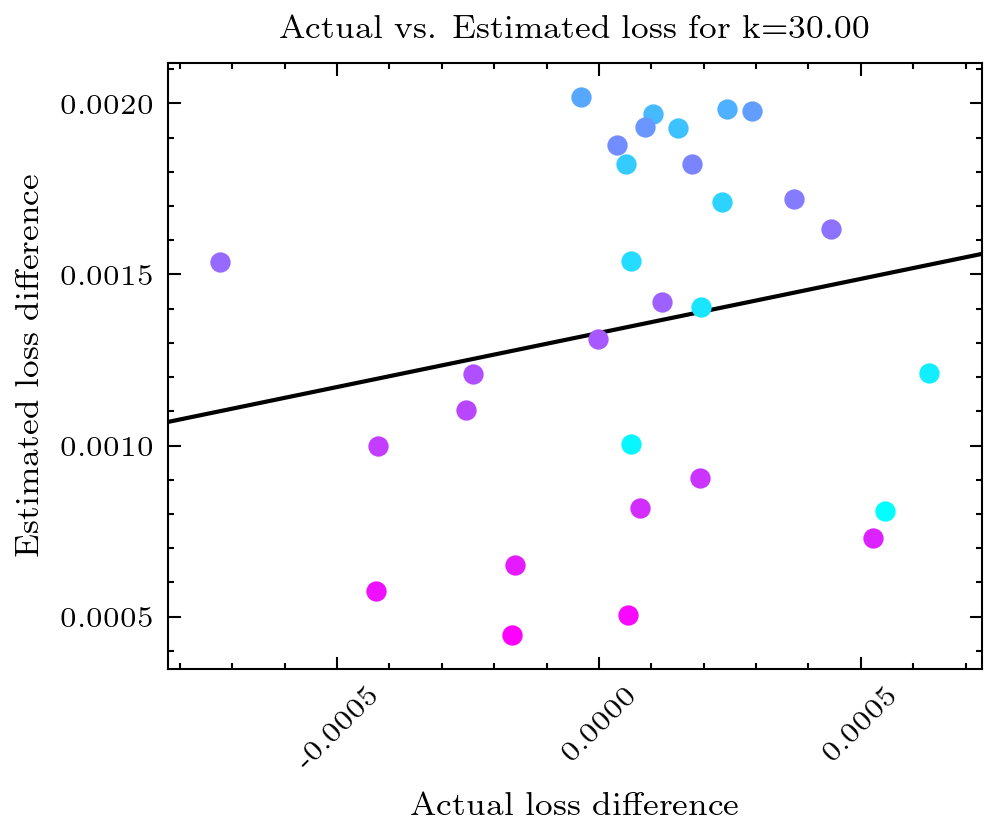}
    \caption{Actual difference in test loss vs. estimated difference in test loss for all $\epsilon$ values per group size for the Adult dataset. Cooler colors depict smaller $\epsilon$ values. Top row: no noise correction, bottom row: with forward loss correction.}
    \label{fig:results}
\end{figure*}
\subsubsection{Time Complexity}
\label{sec:method-time}
\begin{table}[t!]
    \centering
    \caption{Time complexity of $\mathcal{I}_{pert,loss}^{\text{RR}}$ versus retraining for every $\epsilon$ and group under consideration. X: exact, CG: conjugate gradient, SE: stochastic estimation, $n$: number of training points, $p$: number of parameters, $r$: recursion depth for SE estimation, $t$: number of recursions for SE estimation, $E$: number of epochs.}
    \renewcommand{\arraystretch}{1.5}
    \setlength{\tabcolsep}{7pt}
    \begin{tabular}{cccc}
         $\mathcal{I}_{pert,loss}^{\text{RR}}$ (X) & $\mathcal{I}_{pert,loss}^{\text{RR}}$ (CG) & $\mathcal{I}_{pert,loss}^{\text{RR}}$ (SE) & Retrain \\
        \hhline{====}
        $\mathcal{O}(np^2+p^3)$ & $\mathcal{O}(np)$ & $\mathcal{O}(np + rtp)$ & $\mathcal{O}(Enp)$
    \end{tabular}
    \label{tab:complexity}
\end{table}
In Table \ref{tab:complexity}, we detail the computational complexity of calculating the influence function using three different approaches to computing the inverse Hessian vector product (IHVP) as well as the time complexity of normal training of a logistic regression model using gradient based learning. Specifically, we detail the complexity of computing $\mathcal{I}_{pert, loss}^{\text{RR}}$ explicitly and using the conjugate gradient (CG) or stochastic estimation (SE) approaches to estimate the IHVP. We direct interested readers to \cite{koh2017understanding} for a more in depth discussion of how CG and SE can decrease the total computation time. White using the explicit IHVP approach seems to be more computationally complex than retraining, we note that this cost is only accrued once for all $\epsilon$ and groups being considered while retraining would have to be done $e\times g$ times where $e$ is the number of $\epsilon$ values and $g$ is the number of groups being considered. Additionally, using the influence-based approach with CG or SE IHVP estimation can have computational speed-up over the na\"{i}ve retraining approach when the number of training epochs $E$ is large. However we again note that the IHVP only has to be calculated (or estimated using CG/SE) \textit{once} to compute the influence of all $\epsilon$ values ($e$) and group constructions being considered. On the other hand, retraining has to be performed once for every $\epsilon$ and for every group construction. In Section \ref{sec:timing-analysis} we clearly show the power of approximation using influence functions to save computational time even when small models are considered. 

\subsubsection{Using $\mathcal{I}^{\text{RR}}_{pert,loss}$ With Other LDP Protocols} While our analysis and construction has been based on the idea of using randomized response to perturb the features and/or label, here we describe when and how our work can be extended to other LDP protocols such as Randomized Aggregatable Privacy-Preserving Ordinal Response (RAPPOR) \cite{erlingsson2014rappor}, Optimal Linear Hashing (OLH) \cite{wang2017locally}, Optimal Unary Encoding (OUE) \cite{wang2017locally}, Binary Linear Hashing (BLH) \cite{wang2017locally}, and Thresholding with Histogram Encoding (THE) \cite{wang2017locally}. In Eq. \ref{eq:general-inf-rr}, the influence function explicitly considers randomized response as the LDP perturbation scheme due to the scaling term $1-\prod_{f\in\mathcal{F}}\frac{e^{\epsilon_f}}{d_f-1+e^{\epsilon_f}}$. However, this scaling term can be easily altered to represent other perturbation schemes like BLH and OLH. The only restrictions on which LDP perturbation methods can be analyzed using influence functions are: 1) randomization must be done at an individual record level (in order to be able to analyze how changing the record changes the model), and; 2) the the randomization cannot change the feature representation. For example, RAPPOR, OUE, and THE all perform one-hot encoding on the user's input before randomization. This changes the overall feature domain size which makes it complicated to use the randomized value as input to the machine learning model trained on the non-randomized data. In these cases, if the randomized value cannot be fed into the machine learning model trained on non-randomized data (which has a smaller feature domain space), then there is no possible way to see the loss that would be produced on the randomized input. Therefore, the effect of LDP protocols such as OLH and BLH can be approximated using our proposed influence function approach, while other LDP protocols like RAPPOR, OUE, and THE would require modification of our influence based approach since they change the feature domain size.

\section{Evaluation}
\label{sec:eval}
In this section, we present the details of our experimentation to show that the influence functions of Eqs. \ref{eq:general-inf-rr}, \ref{eq:group-to-group}, and \ref{eq:group-to-group-FLC} are able to properly estimate the effect that a certain $\epsilon$ will have on model test loss. Additionally, we provide a detailed timing analysis. We test a standard $\epsilon$ range of 30 evenly spaced $\epsilon$ values from 0.001 to 10. All experiments are run on a Tesla V100 (32GB RAM) GPU. Our code is publicly available at \url{https://tinyurl.com/yc2ra8m4}. 

\subsubsection{Datasets}
We use three datasets in our experimentation: Adult, ACSPublicCoverage (ACSPubCov), and MNIST. Specifically, we perform normal pre-processing (e.g., drop duplicates, perform feature selection, ...), use an 80/20 train/test split, and either binarize (ACSPubCov) or one-hot encode (Adult) the features. This was to aid the influence function in approximating the true loss. Additionally, we chose to use only four classes (1, 3, 7, 8) from the MNIST dataset. This was to allow faster retraining for comparison with our influence based method. Table \ref{tab:datasets} explains the characteristics of the datasets and results of each dataset on a logistic regression model.
\begin{table}[t!]\scriptsize
    \centering
     \caption{Dataset characteristics and the test accuracies using logistic regression. $n$: dataset size after cleaning, $d$: number of original features, OHE-$d$: number of features after cleaning and one-hot encoding, $|c|$: number of classes, Prec: precision, Rec: recall, Acc: accuracy.}
    \label{tab:datasets}
    \setlength{\tabcolsep}{4pt}
    \renewcommand{\arraystretch}{1.3}
    \begin{tabular}{|c|ccccccc|}
        \hline
        Dataset & $n$ & $d$ & OHE-$d$ & $|c|$ & Prec & Rec & Acc \\ \hline
        Adult & 44,355 & 14 & 88 & 2 & 80\% & 74\% & 84\% \\
        ACSPubCov & 52,958 & 19 & 11 & 2 & 71\% & 70\% & 73\% \\
        MNIST & 24,989 & 784 & 784 & 4 & 97\% & 97\% & 97\% \\ \hline
\end{tabular}
\end{table}
\subsubsection{Group Construction}
\label{sec:group_construction}
For the Adult and ACSPubCov dataset we select the group $S$ from the set of training points $Z$ based the \textit{gender} attribute and for the MNIST dataset we randomly select one of the four labels to be the group $S$. During experimentation, we test 10 evenly spaced values of $k$ to use as the group size ranging from $1\%$ to $30\%$ of the selected group $S$. W.L.O.G we set $S_{te} = Z_{te}$. Specifically, we note that in our experimentation we choose to focus on one selected group (e.g., gender) and study how varying this group's size affects the estimation ability of the influence functions. We leave further experimentation over how the selection of the group (e.g., based on age or race) affects the estimation for future work.

\subsubsection{Architecture} Here, we recall that $\mathcal{I}_{pert,loss}^{RR}$ is the general influence function for estimating the effect of applying randomized response-based $\epsilon$-LDP on the features, labels, or features and labels has on the utility of the model (see Eq. \ref{eq:general-inf-rr}), $\mathcal{I}_{pert,loss}^{RR-L}$ is the influence function for estimating the effect of applying randomized response-based $\epsilon$-LDP on the labels only (Eq. \ref{eq:group-to-group}, and $\mathcal{I}_{pert,loss}^{RR-FLC}$ is the influence function for estimating the effect of applying randomized response-based $\epsilon$-LDP with FLC (Eq. \ref{eq:group-to-group-FLC}).
For the experiments on $\mathcal{I}_{pert,loss}^{RR}$ and $\mathcal{I}_{pert,loss}^{RR-L}$ we train the original models using the SGDClassifier offered in the scikit-learn python package. In order to calculate the influence function over the various $\epsilon$ and group sizes, the parameters learned via the SGDClassifier were loaded into a Pytorch model. After calculating the approximate effect using influence functions, the data was perturbed using randomized response and another SGDClassifier was trained over the randomized data. For the experiments on $\mathcal{I}_{pert,loss}^{RR-FLC}$, a Pytorch model was used for both the original and retrained model, the loss function used was the Negative Log Likelihood with log softmax to enable proper calculation of FLC, and we used a learning rate of 0.001 for the Adult and ACSPubCov dataset and 0.5 for MNIST. 

\subsubsection{Metrics}
Based on the analysis performed in \cite{koh2019accuracy}, we use two metrics to evaluate how well influence functions can approximate the true effect $\epsilon$ has on model performance: Spearman's rank correlation coefficient ($\rho$) and mean absolute error (MAE). The value of $\rho$  tells to what degree the estimated effect $\mathcal{I}^{\text{RR}}_{pert,loss}$ and the actual effect $\mathcal{I}_{actual}=|\mathcal{L}(Z_{te},\hat{\theta}_{S,\epsilon})- \mathcal{L}(Z_{te}, \hat{\theta})|$ rank subsets of points similarly, where $\hat{\theta}_{S,\epsilon}$ are the optimal parameters of the model trained over the training set with $S$ that has been perturbed with randomized response parameterized by $\epsilon$. MAE gives measurement of how far apart (on average) our estimated value $\mathcal{I}^{\text{RR}}_{pert,loss}$ and true value $\mathcal{I}_{actual}$ are when we apply $\epsilon$-LDP. We note that we run each experiment ten times and report the average result.

\subsection{Analysis of $\mathcal{I}_{pert,loss}^{RR-L}$}
\begin{table}[t!]\scriptsize
\caption{Mean Absolute Error (MAE) and Correlation coefficient ($\rho$) for all group sizes. For each group size $k$, $\rho$ is computed over all $\epsilon$ values.}
\label{tab:mae-rho}
\center
\setlength{\tabcolsep}{4pt}
\renewcommand{\arraystretch}{1.3}
\begin{tabular}{cc|cc|cc|cc|}
\cline{3-8}
 &  & \multicolumn{2}{c|}{Adult} & \multicolumn{2}{c|}{ACSPubCov} & \multicolumn{2}{c|}{MNIST} \\ \cline{2-8} 
\multicolumn{1}{c|}{} & $k\%$ & \multicolumn{1}{c|}{$\mathcal{I}^{\text{RR-L}}$} & $\mathcal{I}^{\text{RR-FLC}}$ & \multicolumn{1}{c|}{$\mathcal{I}^{\text{RR-L}}$} & $\mathcal{I}^{\text{RR-FLC}}$ & \multicolumn{1}{c|}{$\mathcal{I}^{\text{RR-L}}$} & $\mathcal{I}^{\text{RR-FLC}}$ \\ \hline
\multicolumn{1}{|c|}{\multirow{10}{*}{MAE}} & 1.00 & 0.0005 & 0.0003 & 0.0002 & 0.0009 & 0.0009 & 0.0002 \\
\multicolumn{1}{|c|}{} & 4.22 & 0.0017 & 0.0015 & 0.0015 & 0.0009 & 0.0027 & 0.0007 \\
\multicolumn{1}{|c|}{} & 7.44 & 0.0038 & 0.0004 & 0.0015 & 0.0009 & 0.0031 & 0.0026 \\
\multicolumn{1}{|c|}{} & 10.67 & 0.0065 & 0.0003 & 0.0012 & 0.0007 & 0.0036 & 0.0023 \\
\multicolumn{1}{|c|}{} & 13.89 & 0.0098 & 0.0014 & 0.0012 & 0.0007 & 0.0082 & 0.0020 \\
\multicolumn{1}{|c|}{} & 17.11 & 0.0132 & 0.0009 & 0.0012 & 0.0008 & 0.0082 & 0.0014 \\
\multicolumn{1}{|c|}{} & 20.33 & 0.0163 & 0.0005 & 0.0016 & 0.0008 & 0.0103 & 0.0030 \\
\multicolumn{1}{|c|}{} & 23.56 & 0.0210 & 0.0003 & 0.0022 & 0.0009 & 0.0163 & 0.0021 \\
\multicolumn{1}{|c|}{} & 26.78 & 0.0251 & 0.0012 & 0.0037 & 0.0010 & 0.0111 & 0.0080 \\
\multicolumn{1}{|c|}{} & 30.00 & 0.0292 & 0.0013 & 0.0031 & 0.0011 & 0.0283 & 0.0020 \\ 
\hhline{========}
\multicolumn{1}{|c|}{\multirow{10}{*}{$\rho$}} & 1.00 & 0.424 & 0.377 & 0.086 & 0.251 & 0.300 & 0.024 \\
\multicolumn{1}{|c|}{} & 4.22 & 0.855 & 0.115 & 0.069 & 0.146 & 0.496 & 0.223 \\
\multicolumn{1}{|c|}{} & 7.44 & 0.924 & 0.006 & 0.075 & 0.351 & 0.767 & 0.691 \\
\multicolumn{1}{|c|}{} & 10.67 & 0.952 & 0.149 & 0.506 & 0.051 & 0.746 & 0.700 \\
\multicolumn{1}{|c|}{} & 13.89 & 0.943 & 0.369 & 0.538 & 0.193 & 0.830 & 0.732 \\
\multicolumn{1}{|c|}{} & 17.11 & 0.960 & 0.220 & 0.733 & 0.136 & 0.839 & 0.741 \\
\multicolumn{1}{|c|}{} & 20.33 & 0.969 & 0.067 & 0.738 & 0.056 & 0.889 & 0.846 \\
\multicolumn{1}{|c|}{} & 23.56 & 0.990 & 0.019 & 0.776 & 0.088 & 0.870 & 0.771 \\
\multicolumn{1}{|c|}{} & 26.78 & 0.998 & 0.091 & 0.871 & 0.370 & 0.894 & 0.826 \\
\multicolumn{1}{|c|}{} & 30.00 & 0.992 & 0.243 & 0.835 & 0.099 & 0.838 & 0.691 \\ \hline
\end{tabular}
\end{table}
\label{exp:g2g}
For each $\epsilon$ and group size listed previously, we test to what degree the influence function of Eq. \ref{eq:group-to-group} is able to estimate the true change in model test loss when perturbation via $\epsilon$-LDP and retraining actually occurs. We report the MAE and correlation coefficient $\rho$ values in Table \ref{tab:mae-rho} and graph the results of the Adult dataset on the top row of Fig. \ref{fig:results}. Fig. \ref{fig:results} shows that the Adult dataset has a very strong correlation (average of 0.9) between the actual change in test loss and the approximate change calculated using Eq. \ref{eq:group-to-group}. Having a large correlation coefficient $\rho$ means that the influence function can rank the $\epsilon$ values (according to how they affect the test loss) similarly to how they are ranked when retraining is actually performed. Additionally, the MAE between the true and estimated test loss is relatively low (average of 0.013) especially when the group size is small. For all group sizes, the highest MAE values is 0.029 which is a relatively good estimation of the true change in test loss. The correlation results on the ACSPubCov data are worse than the Adult dataset (especially at small group sizes) with an average $\rho$ of 0.523, but the MAE is better with an average of 0.002. We attribute the worse correlation results on the ACSPublicCoverage dataset to the number of $\epsilon$ values tested. When the $\epsilon$ range was changed to only have 20 values between 0.001 and 5, the average $\rho$ increases to 0.615. The results on the MNIST dataset are also good with an average $\rho$ value of 0.747 as well as good MAE average of 0.009. We note that the lower correlation values on the MNIST dataset are due to using the SE approach to calculate the IHVP for the MNIST dataset whereas we explicitly calculated the inverse Hessian product for the Adult and ACSPubCov datasets as the models were small enough to do so. Overall, the results show that influence functions are able to properly capture the true effect that occurs when $\epsilon$-LDP is applied to the labels of the training dataset.  

\subsection{Analysis of $\mathcal{I}_{pert,loss}^{RR-FLC}$}

\begin{figure*}[t!]
    \centering
    \includegraphics[width=.3\textwidth]{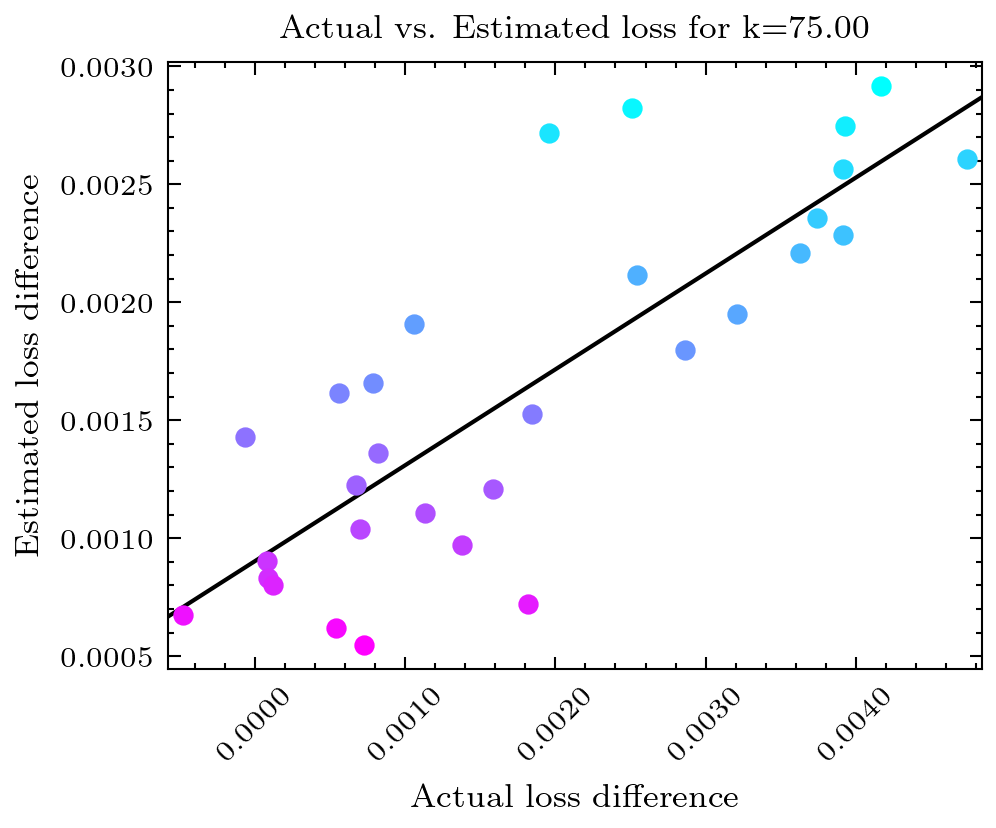}
    \includegraphics[width=.3\textwidth]{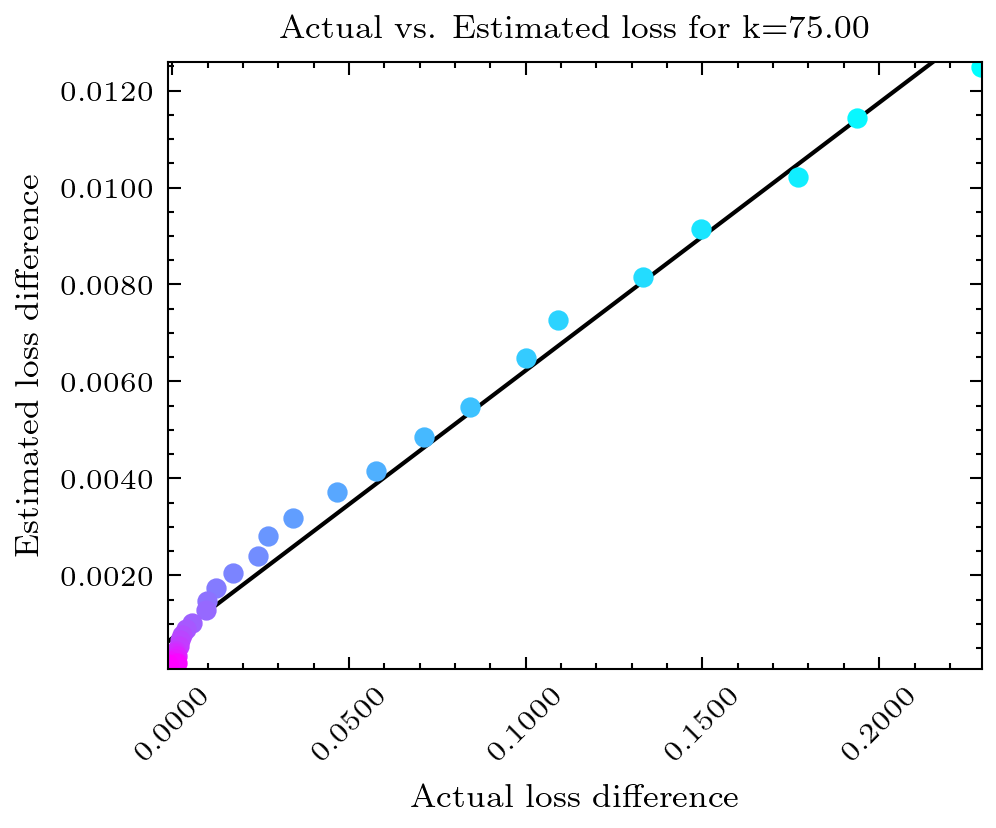}
    \includegraphics[width=.3\textwidth]{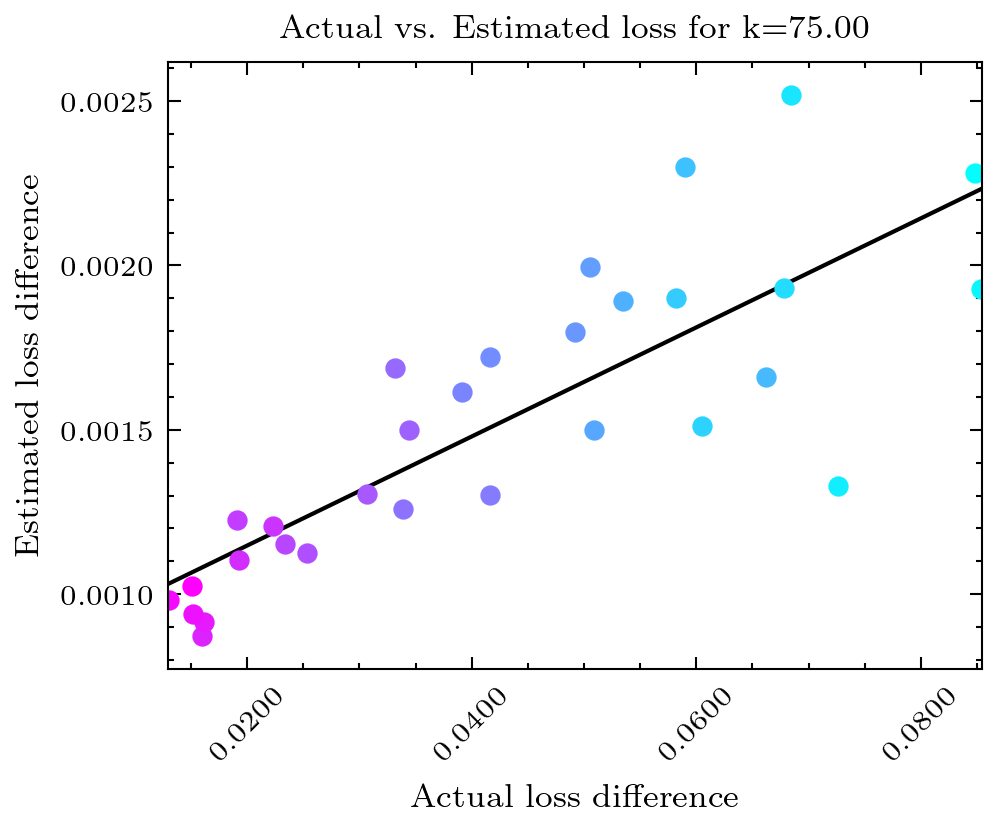}
    \caption{Actual difference in test loss vs. estimated difference in test loss when randomization is performed on two features (left), on the label (middle), and two features plus the label (right) for 75\% of the training set of ACSPublicCoverage. Cooler colors depict smaller $\epsilon$ values.}
    \label{fig:acspubcov-features}
\end{figure*}

Similar to Section \ref{exp:g2g}, we test to what degree the influence function of Eq. \ref{eq:group-to-group-FLC} is able to estimate the true change in model test loss when the perturbation via $\epsilon$-LDP, noise correction, and retraining actually occurs. We report all results in Table \ref{tab:mae-rho} and plot the results for the Adult dataset on the second row of Fig. \ref{fig:results}. For all datasets, the MAE is significantly smaller in comparison to $\mathcal{I}_{pert,loss}^{\text{RR-L}}$. For example, when $k=17.11\%$, the MAE on the Adult dataset reduces from 0.0132 to 0.0009 (a $\sim$93\% decrease), from 0.0012 to 0.0008 on the ACSPublicCoverage dataset, and from 0.0082 to 0.0014 on the MNIST dataset. However, $\rho$ degrades as well. This decrease in $\rho$ after performing FLC is not surprising since by applying FLC we effectively remove the perturbations caused by $\epsilon$-LDP. However, $\mathcal{I}_{pert,loss}^{\text{RR-FLC}}$ is still able to accurately predict the change in test loss with very small MAE meaning that it is still a good approximation method to actually performing retraining.

\subsection{Analysis of $\mathcal{I}_{pert,loss}^{RR}$}
While most of the experimental evaluation has been performed on $\mathcal{I}_{pert,loss}^{\text{RR-L}}$ and $\mathcal{I}_{pert,loss}^{\text{RR-FLC}}$ to show that influence function approximation works in cases both with and without noise correction, here we briefly analyze the ability of our general influence equation in Eq. \ref{eq:general-inf-rr} to approximate the change in test loss when randomized response is performed on the features only, the labels only, or on the features and labels. We show our results on the ACSPublicCoverage dataset for a group size $k=75\%$ in Fig. \ref{fig:acspubcov-features}. When randomized response is performed on the features only, our influence based method is able to approximate the true change that would occur with an average MAE of 0.0009 and a $\rho$ of $0.797$. When randomized response is performed on the labels only, our influence based method is able to approximate the true change that would occur with an average MAE of 0.0468 and a $\rho$ of 0.991. Finally, when randomized response is performed on both the features and the labels, our influence function based method is able to approximate the true change with an average MAE of 0.0407 and a $\rho$ of 0.859. All these results point towards our formulated influence function being a good approximator for the true change in test loss.
\begin{table}[t!]\scriptsize
\centering
\caption{Time for $\mathcal{I}_{pert,loss}^{\text{RR-L}}$ when explicitly calculating the IHVP vs. Retraining.}
\label{tab:time}
\setlength{\tabcolsep}{4pt}
\renewcommand{\arraystretch}{1.5}
\begin{tabular}{c|c|c|}
\cline{2-3}
 & $\mathcal{I}_{pert,loss}^{\text{RR-L}}$ & Retraining \\ \hline
\multicolumn{1}{|c|}{Initial Training} & 5.19 s & 5.19 s \\ \hline
\multicolumn{1}{|c|}{IHVP} & 0.191 s & - \\ \hline
\multicolumn{1}{|c|}{Per $\epsilon$} & Compute IHVP$\cdot\nabla_\theta$ : 0.002 s & Retrain: 3.15 s \\ \hline
\multicolumn{1}{|c|}{Per group} & 0.06 s & 94.5 s \\ \hline
\multicolumn{1}{|c|}{For 10 groups} & 5.19 s + 0.191s + 0.6 s = 5.98 s & 5.19 s + 945 s = 950.19 s \\ \hline
\end{tabular}
\end{table}
\subsection{Time Analysis}
\label{sec:timing-analysis}
To calculate all the estimated changes in test loss using the influence function approach, we have computational cost incurred by three main sources: 1) training the original model on the clean data to get the optimal parameters, 2) calculating the inverse Hessian vector product (IHVP: $-\nabla_{\theta}\mathcal{L}(S_{te},\hat{\theta})^TH_{\hat{\theta}}^{-1}$), and 3) calculating the influence function for each $\epsilon$ per each group size. Additional computational costs will be incurred if model retraining is performed to fit a graph like those shown in Fig. \ref{fig:results}. In Table \ref{tab:time}, we report the average time (out of three runs) to calculate $\mathcal{I}_{pert,loss}^{\text{RR-L}}$ using an \textit{explicit} calculation of IHVP and the average time to perform retraining for all different $\epsilon$ values and group sizes. The first row shows the time to train the original model on the clean data (which is the same for both $\mathcal{I}_{pert,loss}^{\text{RR-L}}$ and retraining). The second row shows the time to compute the IHVP (which, we note, only has to be performed \textit{once} regardless of how many $\epsilon$ values or group sizes are tested). The third row shows the average time to compute the approximate or true change for one $\epsilon$ value and group size. The fourth row shows the average time to compute the approximate or real change for a single group size. In our experimentation, we tested 30 $\epsilon$ values per group size. Finally, the last row shows the average time to compute the approximate or real change for all 10 wanted group sizes, each of which consider all 30 $\epsilon$ values. It is clear that using an influence based approach saves substantial time and computational resources even on small models like those used in the experimentation. We note, however, that the time to calculate $\mathcal{I}_{pert,loss}^{\text{RR-L}}$ should not change significantly even if larger or deeper models are used. This is because when calculating the influence over a neural network or CNN model, normally only the weights of the last fully connected layer are used in the calculation \cite{koh2017understanding}. On the other hand, a deeper or larger model would cause the time of model retraining to increase proportional to the size of the model. Further, we note that using an approximation approach for the calculation of the IHVP (such as conjugate gradient or stochastic estimation) would offer additional computational time savings and we leave experimentation over these settings for future work. 

Due to space constraints, our timing analysis focuses on the label perturbation case only. However, we note that the computation time for the general case ($
\mathcal{I}^{\text{RR}}_{pert,loss}$) is similar to that of the labels -- especially in cases when all the features selected to be perturbed have small domains. When they have large feature domains,  the computational cost increases slightly due to the requirement of calculating the potential loss under all the different possible feature combinations (see Eq. \ref{eq:general-inf-rr}). $
\mathcal{I}^{\text{RR-FLC}}_{pert,loss}$ also incurs minor additional costs (that scales with the domain size of the label) due to the required matrix multiplication to correct the loss (i.e., $\mathrm{\textbf{P}}^Th(x^i;\hat{\theta})$). We leave the timing analysis over $
\mathcal{I}^{\text{RR}}_{pert,loss}$/$
\mathcal{I}^{\text{RR-FLC}}_{pert,loss}$ for future work.

\subsection{Discussion}
While in our experimentation we performed retraining of the model for every $\epsilon$ and group size under consideration, this was to show the ability of the influence function to approximate the true change and is not necessary in practice. In cases where noise correction is applied, the MAE is small and the calculated influence itself serves as a good approximation to the true change that would occur in the test loss under randomized response-based $\epsilon$-LDP. In cases where noise correction is not applied, the MAE is larger and not as good of a representation of the true change. However, the correlation between the estimated and true change is strong. This means that we can plot a graph similar to Fig. \ref{fig:results} to derive what the true change in test loss would be when only given the approximate change. To generate such a graph, a few $\epsilon$ values can be selected to perform model retraining for and then a line can be fit to the resulting values. Additionally, in our experimentation, we set $S_{te}=Z_{te}$. However, when the test set is large this can cause the computation of the influence to increase. To shorten the computation time, a random sample can be taken from $Z_{te}$ to be used as $S_{te}$ without significantly affecting the ability of the influence function to approximate the true change in test loss.

\section{Conclusion}
\label{sec:conclusion}
In this work, we propose an approach based on influence functions to estimate the effect that a chosen $\epsilon$ in randomized response-based $\epsilon$-LDP has on model utility loss. We show that our method is able to accurately approximate the true change that would occur if the data (e.g., features, labels, or features and labels) were actually perturbed, in cases both with and without noise correction applied, and the model was retrained. Further, we show that our method can offer significant computational speed-up over the na\"{i}ve retraining approach. Our future work includes considering how the choice of $\epsilon$ affects other metrics such as the final fairness of the model.

\section*{Acknowledgements}
This work was supported  in part by NSF 1920920 and 1946391.

\bibliographystyle{splncs04}
\bibliography{ref.bib}

\begin{thebibliography}{10}
\providecommand{\url}[1]{\texttt{#1}}
\providecommand{\urlprefix}{URL }
\providecommand{\doi}[1]{https://doi.org/#1}

\bibitem{apple-dp}
Apple~Inc., D.P.T.: Learning with privacy at scale  (7)

\bibitem{bae2020influence}
Bae, J., Ng, N., Lo, A., Ghassemi, M., Grosse, R.: If influence functions are the answer, then what is the question? arXiv:2209.05364  (2022)

\bibitem{10.1145/2746539.2746632}
Bassily, R., Smith, A.: Local, private, efficient protocols for succinct histograms. In: 47th ACM STOC. p. 127–135. NY, USA (2015)

\bibitem{basu2020second}
Basu, S., You, X., Feizi, S.: On second-order group influence functions for black-box predictions. In: ICML. pp. 715--724. PMLR (2020)

\bibitem{bebensee2019local}
Bebensee, B.: Local differential privacy: a tutorial. arXiv:1907.11908  (2019)

\bibitem{chaudhuri2020randomized}
Chaudhuri, A., Mukerjee, R.: Randomized response: Theory and techniques. Routledge (2020)

\bibitem{cook1986assessment}
Cook, R.D.: Assessment of local influence. Journal of the Royal Statistical Society: Series B (Methodological)  \textbf{48}(2),  133--155 (1986)

\bibitem{cook1982residuals}
Cook, R.D., Weisberg, S.: Residuals and influence in regression. New York: Chapman and Hall (1982)

\bibitem{deng2020interpreting}
Deng, Z., Dwork, C., Wang, J., Zhang, L.: Interpreting robust optimization via adversarial influence functions. In: ICML. pp. 2464--2473. PMLR (2020)

\bibitem{ding2017collecting}
Ding, B., Kulkarni, J., Yekhanin, S.: Collecting telemetry data privately. NeurIPS  \textbf{30} (2017)

\bibitem{dwork2006differential}
Dwork, C.: Differential privacy. In: 33rd International Colloquium, ICALP, Venice, Italy, Part II 33. pp. 1--12. Springer (2006)

\bibitem{dwork2009adifferential}
Dwork, C.: The differential privacy frontier. In: 6th TCC. pp. 496--502. Springer (2009)

\bibitem{dwork2019differential}
Dwork, C., Kohli, N., Mulligan, D.: Differential privacy in practice: Expose your epsilons! Journal of Privacy and Confidentiality  \textbf{9}(2) (2019)

\bibitem{dwork2009bdifferential}
Dwork, C., Lei, J.: Differential privacy and robust statistics. In: 41st ACM STOC. pp. 371--380 (2009)

\bibitem{erlingsson2014rappor}
Erlingsson, {\'U}., Pihur, V., Korolova, A.: Rappor: Randomized aggregatable privacy-preserving ordinal response. In: ACM SIGSAC. pp. 1054--1067 (2014)

\bibitem{garfinkel2018issues}
Garfinkel, S.L., Abowd, J.M., Powazek, S.: Issues encountered deploying differential privacy. In: Proceedings of the 2018 Workshop on Privacy in the Electronic Society. pp. 133--137 (2018)

\bibitem{geyer2017differentially}
Geyer, R.C., Klein, T., Nabi, M.: Differentially private federated learning: A client level perspective. arXiv:1712.07557  (2017)

\bibitem{hsu2014differential}
Hsu, J., Gaboardi, M., Haeberlen, A., Khanna, S., Narayan, A., Pierce, B.C., Roth, A.: Differential privacy: An economic method for choosing epsilon. In: 27th CSF. pp. 398--410. IEEE (2014)

\bibitem{kang2020differentially}
Kang, Y., Liu, Y., Ding, L., Liu, X., Tong, X., Wang, W.: Differentially private erm based on data perturbation. arXiv:2002.08578  (2020)

\bibitem{kasiviswanathan2011can}
Kasiviswanathan, S.P., Lee, H.K., Nissim, K., Raskhodnikova, S., Smith, A.: What can we learn privately? SICOMP  \textbf{40}(3),  793--826 (2011)

\bibitem{koh2017understanding}
Koh, P.W., Liang, P.: Understanding black-box predictions via influence functions. In: ICML. pp. 1885--1894. PMLR (2017)

\bibitem{koh2019accuracy}
Koh, P.W.W., Ang, K.S., Teo, H., Liang, P.S.: On the accuracy of influence functions for measuring group effects. NeurIPS  \textbf{32} (2019)

\bibitem{lee2011much}
Lee, J., Clifton, C.: How much is enough? choosing $\varepsilon$ for differential privacy. In: 14th ISC. pp. 325--340. Springer (2011)

\bibitem{li2022achieving}
Li, P., Liu, H.: Achieving fairness at no utility cost via data reweighing with influence. In: ICML. pp. 12917--12930. PMLR (2022)

\bibitem{mehner2021towards}
Mehner, L., von Voigt, S.N., Tschorsch, F.: Towards explaining epsilon: A worst-case study of differential privacy risks. In: EuroS\&PW. pp. 328--331. IEEE (2021)

\bibitem{patrini2017making}
Patrini, G., Rozza, A., Krishna~Menon, A., Nock, R., Qu, L.: Making deep neural networks robust to label noise: A loss correction approach. In: CVPR. pp. 1944--1952 (2017)

\bibitem{pruthi2020estimating}
Pruthi, G., Liu, F., Kale, S., Sundararajan, M.: Estimating training data influence by tracing gradient descent. NeurIPS  \textbf{33},  19920--19930 (2020)

\bibitem{reid2010composite}
Reid, M.D., Williamson, R.C.: Composite binary losses. JMLR  \textbf{11},  2387--2422 (2010)

\bibitem{sajadmanesh2021locally}
Sajadmanesh, S., Gatica-Perez, D.: Locally private graph neural networks. In: ACM SIGSAC. pp. 2130--2145 (2021)

\bibitem{wang2022understanding}
Wang, J., Wang, X.E., Liu, Y.: Understanding instance-level impact of fairness constraints. In: ICML. pp. 23114--23130. PMLR (2022)

\bibitem{wang2017locally}
Wang, T., Blocki, J., Li, N., Jha, S.: Locally differentially private protocols for frequency estimation. In: USENIX Security 17. pp. 729--745 (2017)

\bibitem{wang2016using}
Wang, Y., Wu, X., Hu, D.: Using randomized response for differential privacy preserving data collection. In: EDBT/ICDT Workshops. vol.~1558, pp. 0090--6778 (2016)

\bibitem{warner1965randomized}
Warner, S.L.: Randomized response: A survey technique for eliminating evasive answer bias. Journal of the American Statistical Association  \textbf{60}(309),  63--69 (1965)

\bibitem{xiong2020comprehensive}
Xiong, X., Liu, S., Li, D., Cai, Z., Niu, X.: A comprehensive survey on local differential privacy. Security and Communication Networks  \textbf{2020},  1--29 (2020)

\bibitem{xue2021toward}
Xue, Y., Niu, C., Zheng, Z., Tang, S., Lyu, C., Wu, F., Chen, G.: Toward understanding the influence of individual clients in federated learning. In: AAAI. vol.~35, pp. 10560--10567 (2021)

\bibitem{ye2021out}
Ye, H., Xie, C., Liu, Y., Li, Z.: Out-of-distribution generalization analysis via influence function. arXiv:2101.08521  (2021)

\end{thebibliography}

\end{document}